\pdfoutput=1
\documentclass[12pt]{article}
\usepackage{amsmath}
\usepackage{natbib}
\usepackage{microtype}
\usepackage{graphicx}
\usepackage{subfig}
\usepackage[table]{xcolor} 
\usepackage{colortbl} 
\usepackage{bm}
\usepackage{pgf, tikz} 
\usetikzlibrary{arrows,automata,fit}
\usetikzlibrary{shapes,snakes}
\usepackage{bbm} 
\usepackage{amsfonts}
\usepackage{amsthm}
\usepackage{bm}
\usepackage{verbatim}
\newtheorem{definition}{Definition}
\newtheorem{proposition}{Proposition}

\usepackage{algorithm}
\usepackage{algorithmic}
\usepackage{booktabs}
\newtheorem{remark}{Remark}%

\newcommand\independent{\protect\mathpalette{\protect\independenT}{\perp}}
\def\independenT#1#2{\mathrel{\rlap{$#1#2$}\mkern2mu{#1#2}}}

\usepackage{hyperref}

\newcommand{\blind}{1}

\begin{document}

\def\spacingset#1{\renewcommand{\baselinestretch}%
{#1}\small\normalsize} \spacingset{1}


\if1\blind
{
 \title{\bf Structural Learning of Simple Staged Trees}
  \author{Manuele Leonelli \\
School of Human Science and Technology, IE University, Madrid, Spain\\
and\\
Gherardo Varando\\
    Image Processing Laboratory, Universitat de València,  València, Spain
}
  \maketitle
} \fi

\if0\blind
{
  \bigskip
  \bigskip
  \bigskip
  \begin{center}
    {\LARGE\bf Title}
\end{center}
  \medskip
} \fi

\bigskip
\begin{abstract}
Bayesian networks faithfully represent the symmetric conditional independences existing between the components of a random vector. Staged trees are an extension of Bayesian networks for categorical random vectors whose graph represents non-symmetric conditional independences via vertex coloring. However, since they are based on a tree representation of the sample space, the underlying graph becomes cluttered and difficult to visualize as the number of variables increases. Here we introduce the first structural learning algorithms for the class of simple staged trees, entertaining a compact coalescence of the underlying tree from which non-symmetric independences can be easily read. We show that data-learned simple staged trees often outperform Bayesian networks in model fit and illustrate how the coalesced graph is used to identify non-symmetric conditional independences.
\end{abstract}

\noindent%

\spacingset{1.45} 
\section{Introduction}
Bayesian networks (BNs) are the most used statistical graphical model which provide an intuitive as well as efficient representation of multivariate data. Although the structure of the network can be expert-elicited, it is often learned from data using search algorithms which explore different network structures \citep[e.g.][for a review]{Daly2011,Neapolitan2004}. Novel algorithms continue to appear to account for more flexible structures and to improve speed when thousands of variables are investigated \citep[e.g.][]{Tsagris2020,Yang2019,Wang2020}. 

One drawback of BNs is that they can only explicitly represent standard, symmetric conditional independence statements. However, in practice, conditional independences may faithfully hold only for specific instantiations of the conditioning variables \citep{Shen2020}. Such independences are usually referred to as context-specific conditional independences or more generally as partial or local conditional independences \citep{Pensar2016}.

For this reason, extensions of BNs which formally account for non-symmetric conditional independences  have been proposed~\citep{Boutilier1996,Cano2012,Chickering1997,Desjardins2008,Friedman1996,Geiger1996,Jaeger2006,Hyttinen2018,Pensar2015,Pensar2016,Poole2003,Salmeron2000} and there has been an increasing interest in formalizing the notion of non-symmetric independence \citep{Corander2019,Nicolussi2021,Shen2020,Tikka2019}. With the exception of \citet{Jaeger2006} and \citet{Pensar2015}, all the above-cited models somehow lose the intuitiveness of BNs since they cannot represent all the models' information into a unique graph.

Staged trees \citep{Smith2008,Collazo2018} are a flexible class of statistical graphical models for categorical variables which, starting from an event tree, explicitly and graphically represent any non-symmetric conditional independence by a partitioning of the vertices of the graph. Staged trees have recently gained popularity thanks to the deployment of the \texttt{stagedtree} R package \citep{Carli2021} and to their use for causal reasoning at Google's Deepmind \citep{Genewein2020}.

Despite being able to represent any non-symmetric independences in a unique graph, the visualization of staged trees becomes cluttered and difficult to interpret as the dimensionality of the problem increases. For this reason, approaches to compress the large amount of information about complex dependence structure encoded by the staged tree have been introduced. \citet{Varando2021} defined algorithms to transform a staged tree into a labeled DAG informing about the type of dependence existing between pairs of variables. \citet{Smith2008} devised a coalescence of the vertices of a staged tree giving a more compact graphical representation called chain event graph (CEG), equivalent to the staged tree.

Here we demonstrate that for generic staged trees learned from data the coalescence rule of \citet{Smith2008} merges only a small number of vertices and thus the practical gain of the CEG representation is often very limited. The class of \emph{simple} staged trees \citep{Smith2008} is one such that the coalescence process into a CEG is optimal, in a sense that we formalize below. The focus of this paper is on this specific subclass of staged trees.

Although there are now many different structural learning algorithms for staged trees and CEGs \citep{Barclay2013,new,Carli2021,Collazo2016,Cowell2014,Freeman2011,Silander2013}, to date there are no methods that learn simple staged trees from data. Here, after formalizing the relationship between simple staged trees and BNs, we introduce novel learning algorithms for simple staged trees and illustrate their use in a variety of datasets. Our applications demonstrate that simple staged trees, whilst still often outperforming BNs in terms of model fit and predictive accuracy, provide a concise graphical representation of a wide array of non-symmetric conditional independences. Furthermore, we provide a first contribution discussing the equivalence class of a simple staged tree. In order to assess causal relationships it is fundamental to explore the equivalence class of a model and there has been a growing interest in developing algorithms to perform this task \citep[e.g.][]{Gorgen2018,gorgen2018discovery,leonelli2021context}.


\section{Simple Bayesian Networks}

Before considering staged trees, we need to introduce BNs and a novel class, termed \emph{simple}, for which the underlying graph is constrained.  Let $G=([p],E)$ be a directed acyclic graph (DAG) with vertex set $[p]=\{1,\dots,p\}$ and edge set $E$. Let $\bm{X}=(X_i)_{i\in[p]}$ be categorical random variables with joint mass function $P$ and sample space $\mathbb{X}=\times_{i\in[p]}\mathbb{X}_i$. For $A\subset [p]$, we let $\bm{X}_A=(X_i)_{i\in A}$ and $\bm{x}_A=(x_i)_{i\in A}$ where $\bm{x}_A\in\mathbb{X}_A=\times_{i\in A}\mathbb{X}_i$. We say that $P$ is Markov to $G$ if, for $\bm{x}\in\mathbb{X}$, 
\begin{equation} \label{eq:markov}
P(\bm{x})=\prod_{k\in[p]}P(x_k \mid \bm{x}_{\Pi_k}),
\end{equation}
where $\Pi_k$ is the parent set of $k$ in $G$ and $P(x_k \vert \bm{x}_{\Pi_k})$ is a shorthand for $P(X_k=x_k \vert \bm{X}_{\Pi_k} = \bm{x}_{\Pi_k})$. 

It is customary to label the vertices of a BN so to respect the topological order of $G$, i.e. a linear ordering of $[p]$ for which only pairs $(i,j)$ where $i$ appears before $j$ in the order can be in the edge set. Of course, there can be multiple permutations of $[p]$ which respect the topological order. Henceforth, we assume that $1,2,\dots,p$ is a topological order of $G$.

The ordered Markov condition implies conditional independences of the form
\begin{equation}
\label{ci}
X_i \independent \bm{X}_{[i-1]}\,\vert\, \bm{X}_{\Pi_i}.
\end{equation}
Henceforth, $P$ is assumed to be strictly positive.
\begin{definition}
Let $G$ be a DAG and $P$ Markov to $G$. The \emph{Bayesian network} model (associated to $G$) is 
\[
\mathcal{M}_G = \{P\in\Delta^{\circ}_{\vert\mathbb{X}\vert-1}\,\vert\, P \mbox{ is Markov to } G\}.
\]
where $\Delta^{\circ}_{\vert\mathbb{X}\vert-1}$ is the ($\vert\mathbb{X}\vert-1$)-dimensional open probability simplex.
\end{definition}


In practice, it is often convenient to work with BNs whose DAG is \textit{decomposable}. A DAG $G=([p],E)$ is said to be decomposable if, for every $i\in[p]$, the set $\Pi_i$ is such that there exists a $j\in \Pi_i$ for which $\Pi_i=\Pi_j\cup\{j\}$. For decomposable BNs, probabilistic inference can be performed exactly using fast algorithms, so that often non-decomposable BNs are first transformed into decomposable ones via the moralization process \citep{Darwiche2009}. Decomposable BN models are regular exponential families, differently to generic BNs which are curved \citep{Geiger2001}.


Here we consider a novel class of DAGs.

\begin{definition}
\label{def:simple}
Let $G$ be a DAG and $\pi$ a topological order of $G$. We say that $G$ is \emph{simple} with respect to $\pi$ if for all $i \in[p-1]$, $\Pi_{\pi(i+1)}\subseteq \Pi_{\pi(i)}\cup\{\pi(i)\}$.
Moreover, we simply say that $G$ is simple if there 
exists a topological order $\pi$ of $G$ such that $G$
is simple with respect to $\pi$. 
\end{definition}

\begin{figure}
     \centering
     \includegraphics[scale = 0.8]{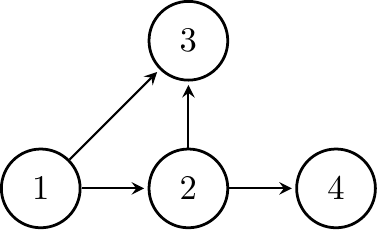}
     \hfill
     \includegraphics[scale = 0.8]{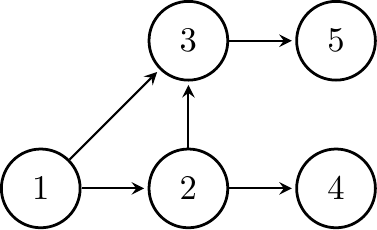}
     \hfill
     \includegraphics[scale=0.8]{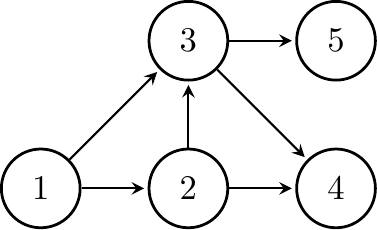}

    \caption{Illustration of simple BNs and the simplification process.}
    \label{fig:simplebn}
\end{figure}

The DAG associated to a total independence model, i.e. having empty edge set, is simple for any permutation of the labelling of the vertices. Similarly, the DAG associated to the saturated model, i.e. there is an edge $(i,j)$ or $(j,i)$ for any two $i,j\in[p]$, is also simple for the only topological order associated to such a DAG. Figure \ref{fig:simplebn} gives further illustrations of the concept of a simple DAG. The DAG in Figure \ref{fig:simplebn} (left) has one topological order ($1234$) and it is simple. The DAG in Figure \ref{fig:simplebn} (center) has two topological orders ($12345$ and $12354$) but for none of these the condition in Definition \ref{def:simple} is met, thus it is not simple. 
 
 The following result is a straightforward consequence of Definition~\ref{def:simple}.

 \begin{proposition}
 \label{prop1}
 Every simple DAG is decomposable.
 \end{proposition}
 
The converse is not true as exemplified by the center DAG in Figure \ref{fig:simplebn}, which is decomposable, but not simple. From Proposition \ref{prop1} it is clear that simple BN models are regular exponential families.
 
Let $1,\dots,p$ be a topological order and fix this order. 
Any DAG $G=([p],E)$ can be transformed into a simple DAG $G'=([p],E')$ by letting the parent sets in $G'$ $\Pi_{i-1}'=\Pi_{i-1}\cup\{\Pi_{i}\setminus\Pi_{i-1}\}$. So, just as in moralization of DAGs, \textit{simplification} transforms the original DAG by adding edges. For the DAG in Figure \ref{fig:simplebn} (middle), the simplification process transforms it into the DAG in Figure \ref{fig:simplebn} (right) by the addition of the edge $(3,4)$.

We refer to Section \ref{sec:simple} for a discussion on how to practically interpret the assumption of simplicity.

\section{Non-Symmetric Models Based on Trees}
Our focus is on models created from trees, differently to BNs which are based on DAGs.

\subsection{X-Compatible Staged Trees}
Consider a $p$-dimensional random vector $\bm{X}$ taking values in the product sample space $\mathbb{X}$. Let $(V,E)$ be a directed, finite, rooted tree with vertex set $V$, root node $v_0$ and edge set $E$. 
For each $v\in V$, 
let $E(v)=\{(v,w)\in E\}$ be the set of edges emanating
from $v$ and $\mathcal{C}$ be a set of labels. 

\begin{definition}
\label{def:x}
An $\bf X$-compatible staged tree 
is a triple $T = (V,E,\theta)$, where $(V,E)$ is a rooted directed tree and:
\begin{enumerate}
    \item $V = {v_0} \cup \bigcup_{i \in [p]} \mathbb{X}_{[i]}$;
		\item For all $v,w\in V$,
$(v,w)\in E$ if and only if $w=\bm{x}_{[i]}\in\mathbb{X}_{[i]}$ and 
			$v = \bm{x}_{[i-1]}$, or $v=v_0$ and $w=x_1$ for some
$x_1\in\mathbb{X}_1$;
\item $\theta:E\rightarrow \mathcal{L}=\mathcal{C}\times \cup_{i\in[p]}\mathbb{X}_i$ is a labelling of the edges such that $\theta(v,\bm{x}_{[i]}) = (\kappa(v), x_i)$ for some 
			function $\kappa: V \to \mathcal{C}$. The function 
			$k$ is called the colouring of the staged tree $T$.
\end{enumerate}
	If $\theta(E(v)) = \theta(E(w))$ then $v$ and $w$ are said to be in the same 	\emph{stage}.
\end{definition} 

Therefore, the equivalence classes induced by  $\theta(E(v))$
form a partition of the internal vertices of the tree  in \emph{stages}.

\begin{figure}
\centering
\includegraphics[]{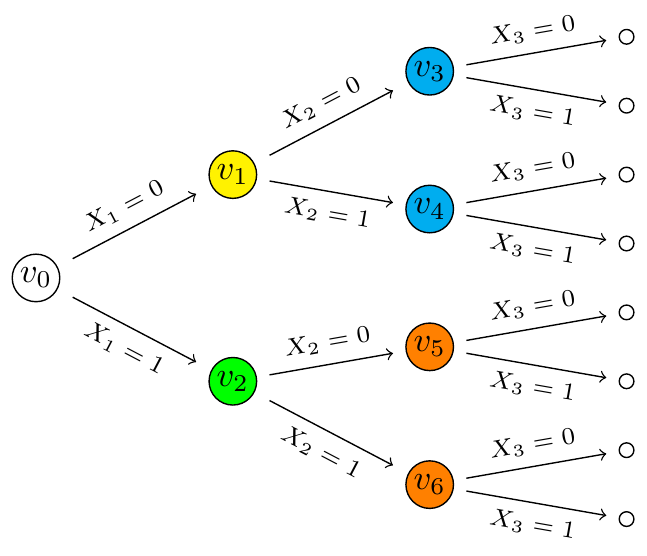}
\caption{An example of an $\bf X$-compatible staged tree. \label{fig:staged1}}
\end{figure}

Definition \ref{def:x} first constructs a rooted tree where each root-to-leaf path, or equivalently each leaf, is associated to an element of the sample space $\mathbb{X}$.  Then a labeling of the edges of such a tree is defined where labels are pairs with one element from a set $\mathcal{C}$ and the other from the sample space $\mathbb{X}_i$ of the corresponding variable $X_i$ in the tree. By construction, $\bf X$-compatible staged trees are such that two vertices can be in the same stage if and only if they correspond to the same sample space. Although staged trees can be more generally defined without imposing this condition \citep[see e.g.][]{Collazo2018}, henceforth, and as common in practice, we focus on $\bf{X}$-compatible staged trees only \citep[see][for an example of a non $\bf{X}$-compatible tree]{Leonelli2019}. 

Figure \ref{fig:staged1} reports an $(X_1,X_2,X_3)$-compatible  staged tree over three binary variables. The \textit{coloring} given by the function $\kappa$ is shown in the vertices and
each edge $(\cdot , (x_1, \ldots, x_{i}))$ is labeled with $X_i = x_{i}$. 
The edge labeling $\theta$ can be read from the graph combining the text label and the 
color of the emanating vertex. 
The staging of the staged tree in Figure \ref{fig:staged1} is given by the partition $\{v_0\}$, $\{v_1\}$, $\{v_2\}$, $\{v_3,v_4\}$ and $\{v_5,v_6\}$.

The parameter space associated to an $\bf X$-compatible staged tree $T = (V, E, \theta)$ 
with 
labeling $\theta:E\rightarrow \mathcal{L}$ 
is defined as
\begin{equation}
\label{eq:parameter}
	\Theta_T=\Big\{\bm{y}\in\mathbb{R}^{\vert \theta(E)\vert} \;\vert \; \forall ~ e\in E, y_{\theta(e)}\in (0,1)\textnormal{ and }\sum_{e\in E(v)}y_{\theta(e)}=1\Big\}.
\end{equation}
Equation~(\ref{eq:parameter}) defines a class of probability mass functions 
over the edges emanating from any internal vertex coinciding with conditional distributions  $P(x_i \vert \bm{x}_{[i-1]})$, $\bm{x}\in\mathbb{X}$ and $i\in[p]$. In the staged tree in Figure \ref{fig:staged1} the staging $\{v_3, v_4\}$ implies that the conditional distribution of $X_3$ given $X_1=0$ and $X_2 = 0$, represented by the edges emanating from $v_3$, is equal to the conditional distribution of $X_3$ given $X_1=0$ and $X_2=1$. A similar interpretation holds for the staging $\{v_5,v_6\}$. This in turn implies that  $X_3\independent X_2\vert X_1$, thus illustrating that the staging of a tree is associated to conditional independence statements.

Let $\bm{l}_{T}$ denote the leaves of a staged tree $T$. Given a vertex $v\in V$, there is a unique path in $T$ from the root $v_0$ to $v$, denoted as $\lambda(v)$. 	The \emph{depth} of a vertex $v\in V$ equals the number of edges in $\lambda(v)$. For any path $\lambda$ in $T$, let $E(\lambda)=\{e\in E: e\in \lambda\}$ denote the set of edges in the path $\lambda$.

\begin{definition}
\label{def:stmodel}
	The \emph{staged tree model} $\mathcal{M}_{T}$ associated to the $\bf X$-compatible staged 
	tree $(V,E,\theta)$ is the image of the map
\begin{equation}
\label{eq:model}
\begin{array}{llll}
\phi_T & : &\Theta_T &\to \Delta_{\vert\bm{l}_T\vert - 1}^{\circ} \\
 &  & \bm{y} &\mapsto \Big(\prod_{e\in E(\lambda(l))}y_{\theta(e)}\Big)_{l\in \bm{l}_T}
\end{array}
\end{equation}
\end{definition}

Therefore, staged trees models are such that atomic probabilities are equal to the product of the edge labels in root-to-leaf paths and coincide with the usual factorization of mass functions via recursive conditioning.

\subsection{Staged Trees and Bayesian Networks}

Although the relationship between BNs and staged trees was already formalized 
by \citet{Smith2008}, a formal procedure to represent a BN as a staged tree has been only recently introduced in \citet{Duarte2020}. 

Assume  $\bm{X}$ is topologically ordered with respect to a DAG
$G$ and consider an $\bf X$-compatible staged tree with vertex set $V$,  
edge set $E$ and labeling $\theta$ defined via the 
coloring $\kappa(\bm{x}_{[i]} ) = \bm{x}_{\Pi_{i}}$ of the vertices. The staged tree $T_G$, with vertex set $V$, edge set $E$ and labeling $\theta$
so constructed, is called \emph{the staged tree model of $G$}. 
Importantly,
$\mathcal{M}_G= \mathcal{M}_{T_G}$, i.e. the two models are exactly the same,
since they entail exactly the same factorization of the joint
probability. Clearly, the staging of $T_G$  represents the
Markov conditions associated to the graph $G$.

\begin{figure}
\centering
\includegraphics[]{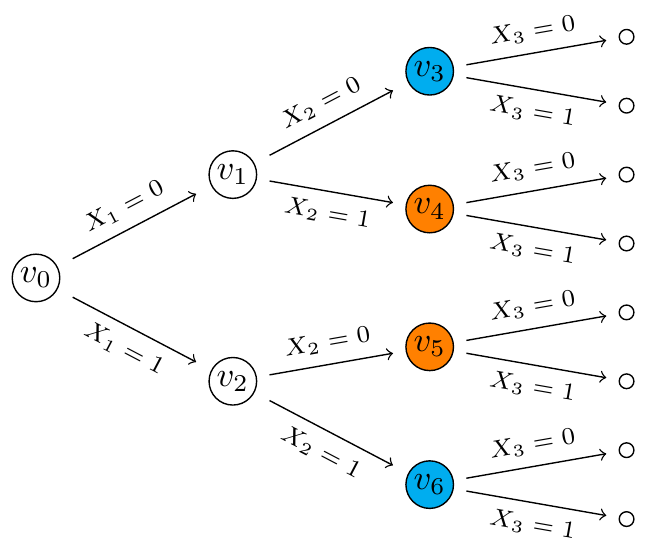}
\caption{An example of a staged tree not associated to any BN $G$. \label{fig:staged2}}
\end{figure}

For instance, the staged tree in Figure \ref{fig:staged1} can be constructed as the $T_G$ from the BN with DAG $X_2\leftarrow X_1\rightarrow X_3$. Conversely, consider the staged tree in Figure \ref{fig:staged2}. The blue staging implies that the conditional distribution of $X_3$ given $X_2=X_1=0$ is equal to the conditional distribution of $X_3$ given $X_2=X_1=1$. Such a constraint cannot be explicitly represented by the DAG of a BN and therefore there is no DAG $G$ such that $\mathcal{M}_G=\mathcal{M}_{T_G}$, i.e. there is no BN which is equivalent to the staged tree in Figure \ref{fig:staged2}.

More generally, \citet{Smith2008} demonstrated that any BN can be represented as an equivalent staged tree, whilst the converse is not true (as in Figure \ref{fig:staged2}).

\subsection{Chain Event Graphs}
Although extremely rich since they can represent any non-symmetric conditional independence (see e.g. Figure \ref{fig:staged2}), the class of staged tree models becomes difficult to visualize as the size of the tree increases. \citet{Smith2008} devised a transformation of a staged tree which, whilst not changing the statistical model, reduces (in theory) the number of vertices and edges of the graph.

Let $T_v$ be the subtree of the $(X_1,\dots,X_p)$-compatible staged tree $T$ rooted at $v\in V$ and suppose that $v$ has depth $k$. Then $T_v$ is a $(X_k,\dots,X_p)$-compatible staged tree with labeling induced by $\theta$. Two internal vertices $v,w\in V$ are said to be in the same \emph{position} if $T_v=T_w$. Of course two vertices can be in the same position only if they are in the same stage. Positions give a coarser partition of the vertex set than stages, i.e. the number of positions is bigger or equal to the number of stages. All leaves of the staged tree are trivially merged in the same position, denoted $w_{\infty}$.

\begin{definition}
Let $T=(V,E,\theta)$ be an $\bf X$-compatible staged tree and $w_1,\dots,w_n,w_{\infty}$ its positions. The associated \emph{chain event graph} $C_T=(V_C,E_C, \kappa_C)$ is a multi-edge graph $(V_C,E_C)$ together with a coloring of the vertices $\kappa_C$,
such that  $V_C=\{w_1,\dots,w_n,w_{\infty}\}$ and the edges $E_C$
are constructed as follows: for every edge  $(v,v') \in E$ 
let $(w, w', x ) \in E_C$ be a labelled edge in the CEG, where $w$ and $w'$ are the positions of $v$ and $v'$ and $\theta(v,v') = (\kappa(v), x)$. 
The coloring $\kappa_C$ over $V_C$ is simply the 
one induced by the coloring $\kappa$ defining $\theta$.
\end{definition}

\begin{figure}
\centering
\includegraphics[]{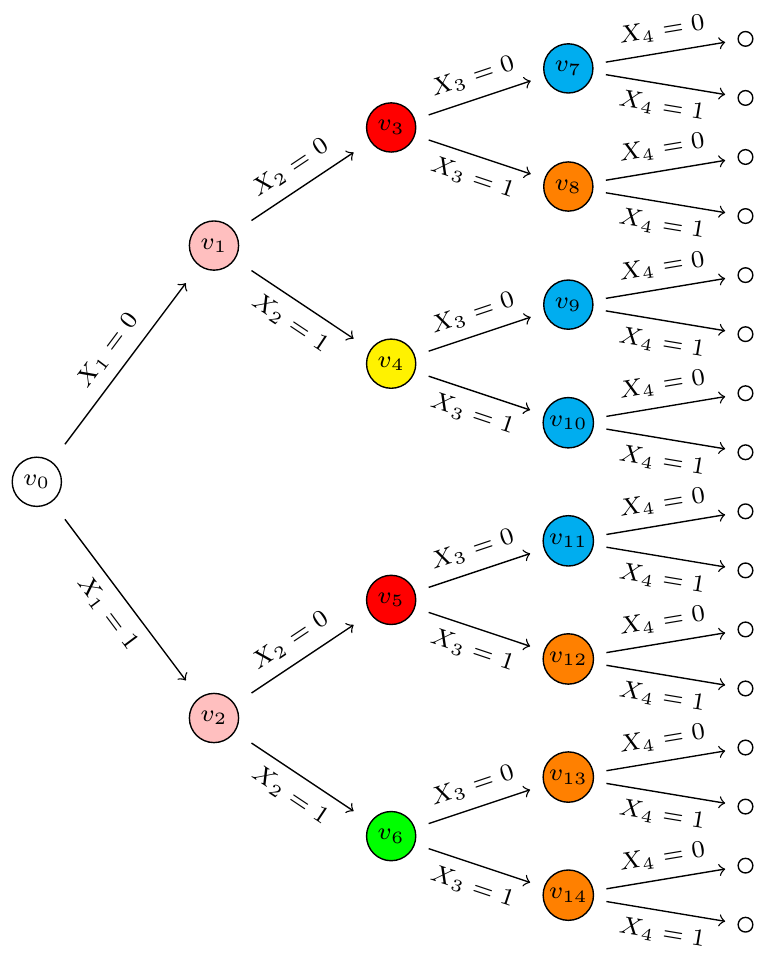}
\caption{Example of an $\bf X$-compatible  staged tree over four binary random variables. \label{fig:staged3}}
\end{figure}

In a CEG all vertices of the staged tree in the same position are merged in a unique vertex. Furthermore, pairs of edges that connect vertices in the same positions are also merged. Therefore the number of vertices and edges of the CEG is smaller than that of the equivalent staged tree.

\begin{figure}
\centering
\includegraphics[]{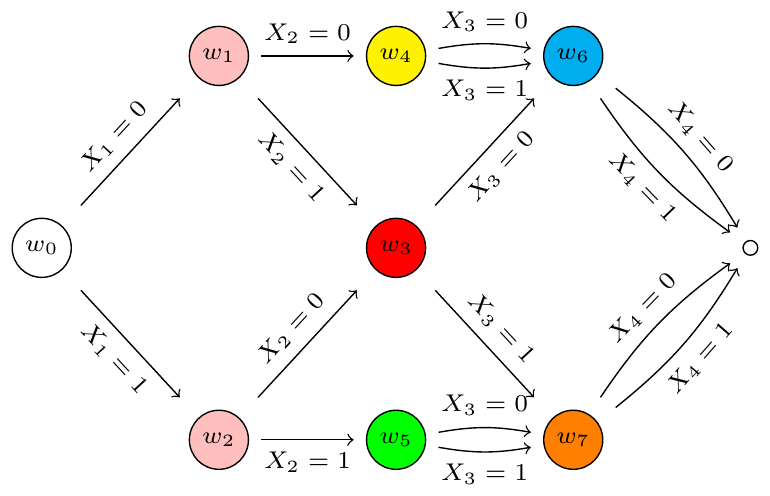}
\caption{The CEG equivalent to the staged tree in Figure \ref{fig:staged3}. \label{fig:ceg}}
\end{figure}

The construction of a CEG is illustrated for the staged tree in Figure \ref{fig:staged3}. The positions of this tree are $w_0=\{v_0\}$, $w_1=\{v_1\}$, $w_2=\{v_2\}$, $w_3=\{v_3,v_5\}$, $w_4=\{v_4\}$, $w_5 = \{v_6\}$, $w_6 = \{v_7,v_9,v_{10},v_{11}\}$ and $w_7 = \{v_8,v_{12},v_{13},v_{14}\}$. These positions together with the leaf $w_{\infty}$ are the vertices of the equivalent CEG reported in Figure \ref{fig:ceg}. The vertices $v_3$ and $v_5$ are in the same position $w_3$ of the staged tree. Their edge labeled $X_3=0$ leads to $v_7$ and $v_{11}$, respectively, which are in the same position $w_6$. Therefore, the two edges are merged in the CEG as can be noticed in Figure \ref{fig:ceg}. Even in this simple example, it is apparent that the CEG is more compact than the associated staged tree, with a reduction from fifteen to eight internal vertices.

It is possible to define the CEG model $\mathcal{M}_{C_T}$ by simply adapting equations (\ref{eq:parameter})-(\ref{eq:model}) to the CEG. Notice that, as in a staged tree, the labeling over the edges emanating from an internal vertex define a probability mass function and root-to-leaf paths are associated to atomic events, whose probabilities can be computed as products of the associated edge labels. Importantly, \citet{Smith2008} demonstrated that $\mathcal{M}_T=\mathcal{M}_{C_T}$.

The CEG does not only provide a more compact graphical representation of the model, but also a framework to perform fast probability propagation \citep{Thwaites2008} and to intuitively read conditional independences from the graph \citep{Thwaites2015}. Illustrations of how to read independences in the CEG are given in Section \ref{sec:app}.

\subsection{Simple Staged Trees}
\label{sec:simple}

Staged trees learned from data are such that often the graphical simplification of transforming the tree into a CEG is minimal. Figure \ref{fig:staged3} gives an idea in this direction since vertices at depth one ($v_1$ and $v_2$), although in the same stage, are not in the same position. In general, vertices close to the root are less likely to be in the same position since the associated subtrees require many symmetries.

\begin{definition}
\label{def:simple1}
An $\bf X$-compatible staged tree $T = (V, E, \theta)$ is \emph{simple} if the vertex partitions into stages and positions coincide. 
That is, if for every vertices $v, w \in V$, 
\[ \theta(E(v)) = \theta(E(w)) \Rightarrow 
T_v = T_w.
\]
\end{definition}

Simple staged trees lead to a very compact CEG since each stage corresponds to exactly one vertex in the CEG. The staged trees in Figures \ref{fig:staged1} and \ref{fig:staged2} are simple, whilst the staged tree in Figure \ref{fig:staged3} is not simple: indeed, its CEG has two vertices belonging to the same stage ($w_1$ and $w_2$).

Definition \ref{def:simple1} gives a graphical condition for a staged tree to be simple, but does not clearly highlight what assumptions underlie simplicity, which is deeply connected to the order $(X_1,\dots,X_p)$ chosen for the variables. 
Fix such an order and assume that two vertices associated to $X_i$ are in the same stage: that is, 
$
P(x_i\vert\bm{x}_{[i-1]})=P(x_i\vert\bm{x}_{[i-1]}'),
$
for all $x_i\in\mathbb{X}_i$ and $\bm{x}_{[i-1]},\bm{x}_{[i-1]}^{'}\in\mathbb{X}_{[i-1]}$. Then in a simple staged tree the previous equality implies, for all $j>i$, that
\begin{equation}
    \label{eq:simple}
P(x_j\vert\bm{x}_{[i-1]},\bm{x}_{[j-1]\setminus[i-1]})=P(x_j\vert\bm{x}_{[i-1]}',\bm{x}_{[j-1]\setminus[i-1]}),
\end{equation}
for all $x_j\in\mathbb{X}_j$ and all $\bm{x}_{[j-1]\setminus[i-1]}\in\mathbb{X}_{[j-1]\setminus[i-1]}$. In words, for all variables $X_j$ following $X_i$ in the assumed order, it must be that that the conditional probability distributions of $X_j$ are equal in the contexts including $\bm{x}_{[i-1]}$ and $\bm{x}_{i-1}'$.

A simple BN, with respect to a topological order $\pi$, is such that the equalities in Equation (\ref{eq:simple}) represent symmetric conditional independences. More generally, in a simple staged trees these may represent non-symmetric independence statements.

\begin{figure}
\centering
\includegraphics[]{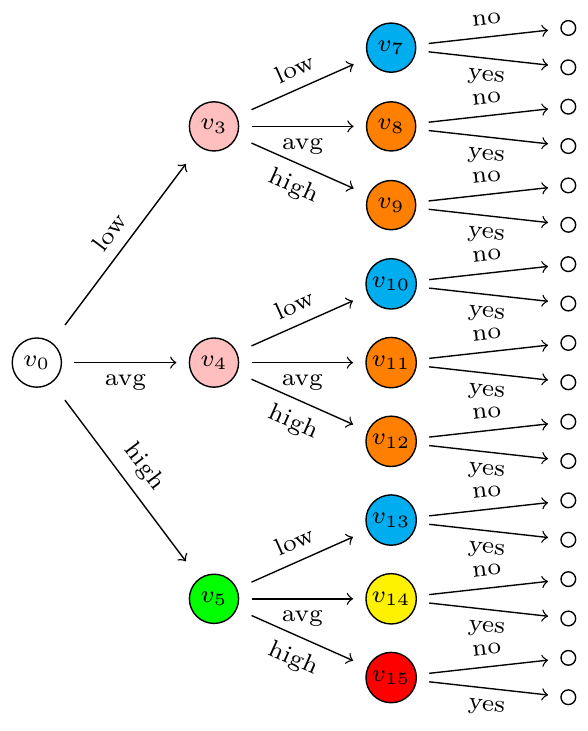}
\caption{A simple staged tree for the Christchurch Health and Development Study. \label{fig:simplex}}
\end{figure}

To illustrate the assumption of simplicity, consider the staged tree in Figure \ref{fig:simplex} inspired by Christchurch Health and Development Study \citep{fergusson2001christchurch}. This studies the effect the family's social background, the economic status and the number of family life events have on a child's health which is measured by rates of hospital admission. We consider three variables: $X_1$ - social and economic situation (low/avg/high); $X_2$ - number of family life events (low/avg/high); $X_3$ - hospital admission (no/yes). An $(X_1,X_2,X_3)$-compatible staged tree is reported in Figure \ref{fig:simplex}. It can be seen that the vertices $v_3$ and $v_4$ are in the same stage, i.e. the distribution of family life events is the same for low and average social and economic backgrounds. Any simple staged tree with staging $\{v_3,v_4\}$ must then be such that $v_7$ and $v_{10}$, $v_8$ and $v_{11}$, and $v_9$ and $v_{12}$ are in the same stage. In words, assuming that the distribution of family life events is the same for low and average social and economic backgrounds implies, for instance, that the distribution of hospital admission is the same for $(X_1 = low, X_2 = low)$ and $(X_1 = avg, X_2 = low)$. It can be seen that the staged tree in Figure \ref{fig:simplex} is indeed simple.

The following proposition, which formalizes a result first appeared in \citet{Smith2008}, establishes the relationship between BNs and simple staged trees.
\begin{proposition}
\label{theo}
Let $T_G$ be the staged tree equivalent to a BN with DAG $G$. Then $T_G$ is simple if and only if $G$ is simple.
\end{proposition}

The result follows from Theorem 3 of \citet{Smith2008}. 
So in particular if $T_G$ is simple then $G$ is also decomposable, but the converse is not true. Another consequence is that if a staged tree is simple then it is either equivalent to a simple BN or it is not equivalent to any BN. For example, the simple staged tree in Figure \ref{fig:staged1} is equivalent to the BN $2 \leftarrow 1 \rightarrow 3$, whilst the simple staged tree in Figure \ref{fig:staged2} is not equivalent to any $G$.

A result similar in spirit to Proposition \ref{theo} was derived in \citet{Duarte2020} formalizing the equivalence between decomposable BNs and \emph{balanced} staged trees, which are characterized by a polynomial condition based on interpolating polynomials \citep{Gorgen2015}, holding for all simple staged trees. \citet{Gorgen2020} demonstrated that balanced staged trees, and thus also simple ones, are regular exponential families. Therefore, the BIC is an asymptotically valid rule for model selection.


\subsection{Structural Learning of Simple Staged Trees}
\label{sec:structalg}
We next discuss the first structural learning algorithms to learn simple staged trees from data. They can be classified into three types: (i) by learning a generic staged tree from data and then simplifying the staged tree (in a similar fashion to simplification of BNs); (ii) directly learning a simple staged tree from data given an ordering of the variables; (iii) directly learning a simple staged tree from data but without assuming a variable ordering a priori.




To \textit{simplify} a staged tree, as described in \cite{Collazo2018},
it is informally sufficient to use positions as stages. More precisely, 
given an $\mathbf{X}$-compatible staged tree $T = (V, E, \theta)$ 
where $\theta((v, \mathbf{x}_{[i]})) = (\kappa(v), x_i)$, let  
$T^s = (V, E, \theta^s)$ be the staged tree obtained by $\theta^s(v, \mathbf{x}_{[i]}) = (w(v), \mathbf{x}_i)$, 
where for each $v \in V$, $w(v)$ is the position of $v$ in 
$T$.
It is easy to see that $T^s$ is a simple staged tree.
The labelling $\theta^s$ is such that the resulting model 
$\mathcal{M}_{T^s}$ is the smallest staged tree model that includes
$\mathcal{M}_T$ among the possible staged trees $(V,E,\theta')$.
Given the above, we can define a first strategy to learn a 
simple $\bf X$-compatible staged tree from data: (1) Use a learning
algorithm to estimate an $X$-compatible staged tree $T$ (for instance, any algorithm of the \texttt{stagedtrees} R package); (2) 
simplify $T$ into $T^s$. 
Such learning strategy has the advantage of being extremely 
simple to implement and it does not add a significant computational 
complexity given that the position identification can be done 
very efficiently~\citep{shenvi20a}. Nevertheless, we will show
that, as main drawback, it outputs CEG with a lot of positions 
rendering them intractable and not easily interpretable.

It is intuitively more efficient to learn directly a simple staged tree from data. We propose two greedy methods for this task which assume a fixed variable ordering.
The first simple algorithm is a hill-climbing 
joining of stages, imposing, for each depth, that stages coincide with positions:
    (1) Start from the full tree where each node 
    is in a different stage and set $i=1$.
    (2) Run a hill-climbing joining 
    of stages for nodes at depth $i$ to 
    optimize the BIC score.
    (3) Impose the stages at level $i$ to be 
          positions by joining relevant stages in all
          the sub-trees (it is sufficient to join stages at depth $i+1$).
          Set $i = i+1$ and repeat from (2) until the whole tree is
          learned.
We name this first method {marginal} algorithm since the 
BIC score is optimized for the (marginal) log-likelihood at a 
specific depth (i.e. for a specific variable) disregarding the effect of joining the stages 
for the sub-trees (the subsequent variables in the ordering). 

The second fixed variable order algorithm corresponds to a hill-climbing joining of 
positions to optimize the BIC score, we name this algorithm 
{total} in contrast with the former approach: 
     (1) start from the full tree where each node 
    is in a different stage and set $i=1$.
    (2) Run a hill-climbing joining 
    of positions for nodes at depth $i$ to 
    optimize the BIC score.
    (3) Set $i = i+1$ and repeat from previous step until all tree is
          learned.
The marginal algorithm is computationally faster since the 
BIC score can be decomposed across the depth of the tree and 
thus, at each step of the search, minimal operations are needed 
to compute the updated score. 
Conversely, the total algorithm is more expensive but the optimized
score is the actual BIC of the model.

If the variables' ordering is 
not known or it cannot be assumed 
beforehand, it is possible 
to combine the previous marginal algorithm
with a greedy order search.
Specifically, the {marginal} 
algorithm can be coupled with 
the greedy approach for order search
implemented in the \texttt{stagedtrees} R package~\citep{Carli2021}.
The greedy method works by 
building the tree one variable at a time. Starting from the root and
adding, at each step, the 
variable which best improves the model score (e.g. BIC). We denote
this algorithm combination
greedy marginal.
Conversely, the {total} 
approach cannot be coupled with the 
greedy approach for order search, 
described previously, since 
the {total} optimization 
for stages/positions at a certain depth affects also 
nodes at subsequent depths.  
Thus, for the {total} 
method we resort to 
perform an exhaustive search over the $p!$ possible variables
orderings.
Obviously such exhaustive search over all possible orders is feasible only for relatively small 
problems ($p\leq 7$ in our computational study of Section~\ref{sec:exp}).

A simple implementation of all the algorithms based on the \texttt{stagedtrees} R package~\citep{Carli2021} is freely available in the repository 

\begin{center}
\href{gherardovarando/simple\ _stagedtrees}{https://github.com/gherardovarando/simple\_stagedtrees} 
\end{center}

\noindent together with the code to reproduce all the examples, simulations and applications. 

\subsection{Equivalence in Simple Staged Trees}

Determining the equivalence class of a statistical graphical model, such as a BN or a staged tree, is a fundamental task to determine if relationships between variables can be assumed to have a causal meaning \citep{Pearl2009,Shafer1996}. By equivalence class of a staged tree $T$, we refer to all staged trees $T'$ whose models $\mathcal{M}_{T'}=\mathcal{M}_{T}$ and therefore cannot be discriminated from data. Algorithms that explore the equivalence class of a staged tree have been recently introduced \citep{Gorgen2018,gorgen2018discovery}. Such algorithms could be of course used to output all the staged trees equivalent to a given simple staged trees.

Here, we provide a first discussion of the properties of the equivalence class for simple staged trees. 



\begin{remark}
\label{rmrk:one}
The equivalence class of a simple staged tree $T$ does not necessarily consists of simple staged trees only.
\end{remark}
This can be easily seen by considering a DAG $G$ such that it is simple with respect to a topological order, but not with respect to another. Let $T_G$ and $T_{G}'$ be the two staged trees associated to such a DAG and the two orders. Then $\mathcal{M}_{T_G}=\mathcal{M}_{T_G'}$ but one is simple and the other is not. Therefore,  by using structural learning algorithms for simple staged trees, one would often learn models that are not equivalent by considering different variable orderings. 
This differs from the recent result of \citet{Duarte2021} which considers a different subclass of staged trees called CStrees. The equivalence class of a given CStree can only include other CStrees.

While Remark~\ref{rmrk:one} seems a drawback at first, 
it is worth noticing that such a 
negative result introduce an asymmetry between 
different trees in the same 
statistical equivalence class and 
thus it could help 
identifying (possibly causal) orderings
of the variables. 
In other words, by considering 
simple staged trees only we reduce the
sets of equivalent candidates belonging
to the same equivalence class. 

\begin{remark}
Let $T$ and $T'$ be two equivalent staged trees, i.e. $\mathcal{M}_T=\mathcal{M}_{T'}$. Then the simplified staged trees $T^s$ and $T'^s$ may not be equivalent. 
\end{remark}

To see this consider the case of two equivalent staged trees, where one of the two is simple. Then simplification of the simple tree does not change its staging, whilst the simplification of the other does. Therefore the two simplified trees are not equivalent.

\section{Experiments}
\label{sec:exp}

We perform computational experiments to showcase
the use of the proposed simple staged trees learning algorithms: first a simulation study with generated simple staged trees as ground truth, and then  we consider nine different  datasets from the literature on  graphical models. 

\subsection{Simulation Study}

We perform a simulation study to investigate the 
performances of the proposed simple staged trees learning algorithms with fixed variables' order. 

We compare the 
classical backward hill-climbing algorithm~\citep{Carli2021} (BHC) and the simplified staged tree  
against the 
proposed simple staged tree learning algorithms: the marginal and total  approaches described in Section~\ref{sec:structalg}.

We generate
simple staged trees by randomly joining positions 
starting from a full 
$\mathbf{X}$-compatible staged tree, where 
$\mathbf{X} = (X_1, \ldots, X_6)$ with $X_i$ binary variables. 
We consider increasing
joining 
probability levels 
$q = 0.2, 0.5, 0.7$, where $q = 0.2$ ($q=0.7$) corresponds to 
trees with more (less) positions.

For each $q$ and 
number of samples $N = 25,\ldots, 2000$ we generate $100$ trees. 
For each generated simple staged tree we learn trees with the different algorithms from $N$ observations sampled from a joint probability compatible with the given staged tree.

\begin{figure}
    \centering
    \includegraphics[scale=0.8]{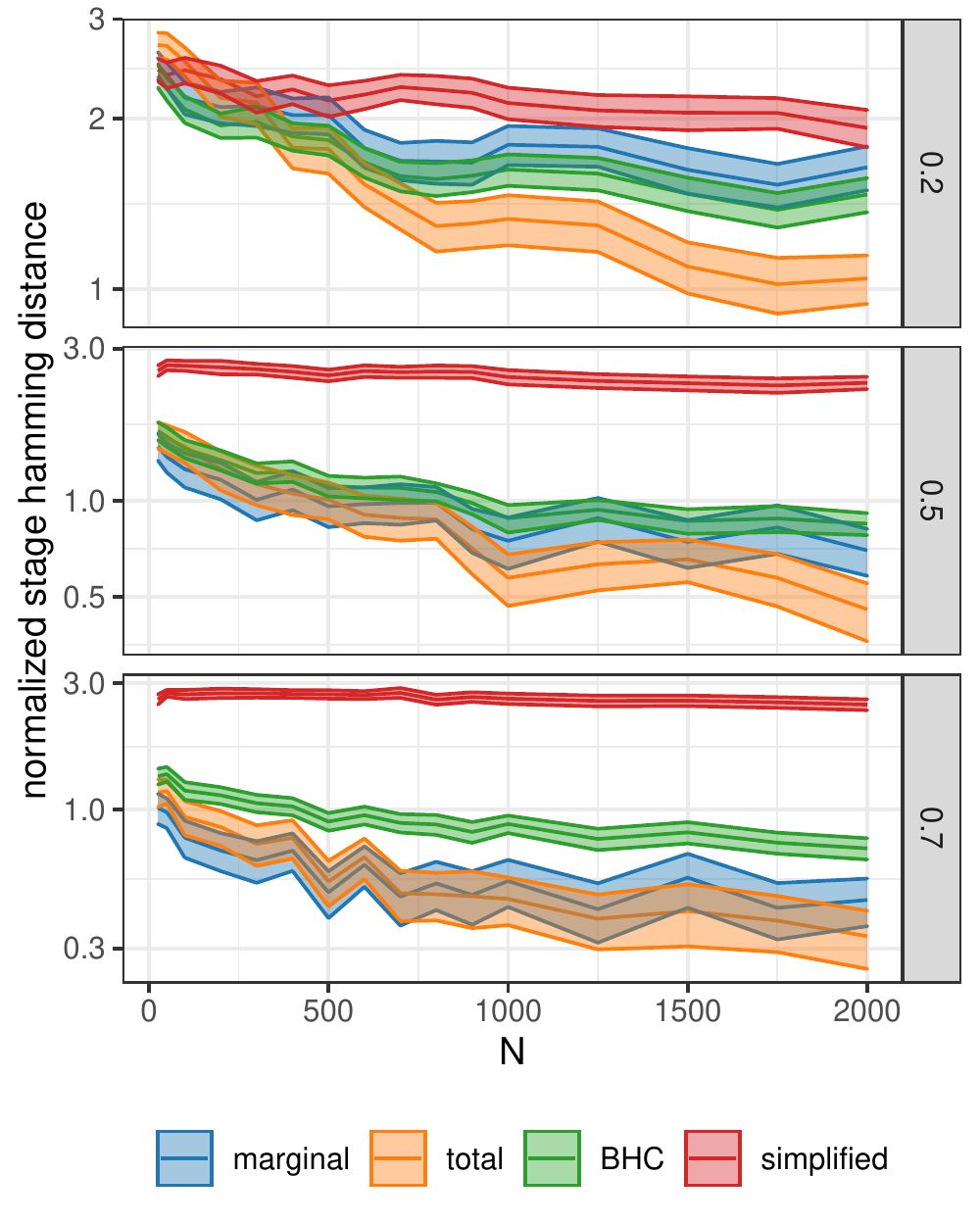}
    \caption{Average results and 95\% confidence intervals for the normalized stage Hamming distances between estimated and true staged trees.}
    \label{fig:simorder}
\end{figure}

We compare the learned staged trees to the true generating tree by computing the \textit{normalized Hamming stage distance}, which we define as the 
sum along the depth of the tree
of the average number of nodes 
whose coloring must be changed to 
obtain the target model.
We report the average and confidence intervals of the normalized Hamming stage distances in Figure~\ref{fig:simorder}.

We notice that, as the sample size increase, the backward hill climbing method, the marginal and the total 
algorithms learn models closer to
the true data generating staged tree.
Conversely, the simplified trees obtained from the BHC estimates are
not converging to the true models. 
We thus deduce that specialized 
simple staged tree learning algorithms are needed to obtain simple models without sacrificing consistency.
As expected, the total algorithm is to be preferred, especially for more
complex models and higher sample sizes. 

\subsection{Computational Study}

\begin{table}[ht]
\centering
\scalebox{0.77}{
\begin{tabular}{lrrrrrrrr}
  \toprule
 & BN & Marg. & Total & BHC & Simpl. & Greedy Marg. & All Marg. & All Total \\ 
  \midrule
asia & 22214.59 & 26225.00 & 22255.09 & 22174.53 & 22350.63 & 22500.89 &  &  \\ 
  cachexia & 963.98 & 984.78 & 1121.77 & 791.98 & 1092.72 & 936.48 &  &  \\ 
  chds & 2831.56 & 2826.09 & 2825.67 & 2824.87 & 2834.70 & 2834.32 & 2825.61 & 2825.61 \\ 
  coronary & 13442.02 & 13972.00 & 13362.36 & 13292.00 & 13369.28 & 13457.04 & 13332.06 & 13322.98 \\ 
  fall & 137807.93 & 138182.64 & 137495.12 & 137442.58 & 137536.66 & 139498.47 & 137468.30 & 137461.77 \\ 
  ksl & 11801.14 & 12194.85 & 12358.51 & 11414.35 & 12319.85 & 12060.13 &  &  \\ 
  mathmarks & 956.03 & 939.16 & 958.50 & 859.05 & 1024.04 & 934.73 & 923.97 & 958.50 \\ 
  phd & 8372.93 & 8787.46 & 8528.46 & 8297.50 & 8526.54 & 8392.60 & 8360.52 & 8405.01 \\ 
  titanic & 10502.28 & 10605.18 & 10450.01 & 10432.84 & 10453.93 & 10502.50 & 10453.93 & 10443.52 \\ 
   \bottomrule
\end{tabular}
}
\caption{BIC for models learned over nine datasets: 
                    Bayesian networks (BN), 
                    simple staged trees with marginal algorithm (Marg.), 
                    simple staged trees with total algorithm (Total), 
                    generic staged trees with backward hill-climbing (BHC), 
                    simple staged trees derived via simplification (Simpl.), 
                    simple staged tree with variable ordering (Greedy Marg.), 
                    simple staged tree considering all orders 
                    and marginal algorithm (All-Marg.), 
                    simple staged tree considering all orders and total algorithm (All-Tot.).} 
\label{table:bic}
\end{table}

Next we consider nine datasets from the literature of probabilistic graphical models to assess the performance of simple staged trees and different estimation procedures. If required, variables are discretized using the equal-frequency method. For each dataset, eight different models are considered. First, a BN model is learned using the hill-climbing algorithm implemented in the \texttt{bnlearn} R package \citep{Scutari2010} (labeled BN). A topological order of the learned BN is then used for learning simple staged trees using both the marginal and the total algorithm (labeled Marg. and Total, respectively). Such an order is also used to learn a more general staged tree using the backward hill-climbing algorithm in the \texttt{stagedtrees} R package (labeled BHC). The resulting staged tree is further simplified to give a simple staged tree (labeled Simpl.). Staged trees without assuming a fixed variable-ordering are also learned using the greedy-marginal algorithm discussed in Section \ref{sec:structalg} (labeled WO-Marg.). Lastly, if computationally feasible, we implemented an exhaustive model search using the marginal and total algorithm for every possible variable ordering (labeled All-Marg. and All-Tot., respectively).

\begin{table}[ht]
\centering
\scalebox{0.77}{
\begin{tabular}{lrrrrrrrr}
  \toprule
 & BN & Marg. & Total & BHC & Simpl. & Greedy Marg. & All Marg. & All Total \\ 
  \midrule
asia & 32 & 12 & 18 & 38 & 38 & 16 &  &  \\ 
  cachexia & 25 & 17 & 10 & 66 & 66 & 17 &  &  \\ 
  chds & 8 & 7 & 7 & 8 & 8 & 6 & 7 & 7 \\ 
  coronary & 21 & 13 & 15 & 27 & 27 & 11 & 14 & 12 \\ 
  fall & 17 & 11 & 14 & 19 & 19 & 10 & 9 & 10 \\ 
  ksl & 37 & 16 & 14 & 168 & 168 & 17 &  &  \\ 
  mathmarks & 13 & 12 & 7 & 38 & 38 & 11 & 10 & 7 \\ 
  phd & 25 & 9 & 10 & 45 & 45 & 12 & 13 & 10 \\ 
  titanic & 21 & 10 & 16 & 17 & 17 & 10 & 11 & 15 \\ 
   \bottomrule
\end{tabular}
}
\caption{Number of positions for the staged trees reported in 
                Table \ref{table:bic}.  For BHC the number in parenthesis is the ratio 
                between the number of stages and the number of positions.} 
\label{table:pos}
\end{table}

The BIC of the learned models is reported in Table \ref{table:bic}. The staged tree learned with the backward hill-climbing has the lowest BIC for all datasets. This is expected since it searches over the largest model space, which includes those of all other models considered. We can further notice that for seven out of the nine datasets (name in bold), one simple staged tree outperforms the BN model, even without considering the algorithms that search all possible variable orderings (except for the cachexia dataset). This highlights that although simple staged trees have a highly constrained topology, they can still provide a good representation of real-world data by embedding non-symmetric conditional independences. Furthermore, one can see that the BIC of models learned with a fixed-variable ordering is not too different to those of models learned without assuming an ordering a priori. 

Although the backward hill-climbing returns the best fitting model, it also learns staged trees whose equivalent CEG has a large number of vertices, as reported in Table \ref{table:pos}. For all datasets, simple staged trees learned directly from data  have a smaller number of positions and therefore the associated CEG representation is much more compact. Table \ref{table:pos} further reports the ratio between the number of stages and the number of positions for the staged tree learned with backward hill-climbing. The lower the ratio, the less effective  the simplification process is (by construction, the ratio for the other trees is one). This highlights the need of algorithms that learn the structure of a simple staged tree from data since otherwise the coalescence of the tree into the CEG merges a very small number of vertices: the resulting CEG would not be any simpler to graphically investigate than the original staged tree.

\begin{table}[ht]
\centering
\scalebox{0.77}{
\begin{tabular}{lrrrrrrrr}
  \toprule
 & BN & Marg. & Total & BHC & Simpl. & Greedy Marg. & All Marg. & All Total \\ 
  \midrule
asia & 0.014 & 0.084 & 0.534 & 0.955 & 0.021 & 0.133 &  &  \\ 
  cachexia & 0.005 & 0.217 & 0.247 & 3.541 & 0.080 & 0.280 &  &  \\ 
  chds & 0.003 & 0.015 & 0.024 & 0.050 & 0.002 & 0.019 & 0.219 & 0.487 \\ 
  coronary & 0.007 & 0.034 & 0.168 & 0.770 & 0.003 & 0.061 & 16.097 & 69.568 \\ 
  fall & 0.022 & 0.033 & 0.473 & 0.105 & 0.003 & 0.074 & 0.896 & 6.022 \\ 
  ksl & 0.010 & 0.115 & 0.149 & 134.996 & 0.052 & 0.156 &  &  \\ 
  mathmarks & 0.002 & 0.060 & 0.049 & 1.761 & 0.006 & 0.093 & 4.347 & 4.400 \\ 
  phd & 0.004 & 0.031 & 0.106 & 3.934 & 0.008 & 0.060 & 21.241 & 39.688 \\ 
  titanic & 0.005 & 0.021 & 0.125 & 0.097 & 0.002 & 0.024 & 0.628 & 1.682 \\ 
   \bottomrule
\end{tabular}
}
\caption{Time (in seconds) to learn the models reported 
                in Table \ref{table:bic}. 
                Computations carried out with an intel CORE i7 vPro 8th Gen.} 
\label{table:time}
\end{table}

Table \ref{table:time} reports the computational times for all models learned. Whilst the BN is the model that requires less time, simple staged trees learned using the marginal and total algorithms require fractions of seconds. Furthermore, they are much faster than the backward-hill climbing algorithm for generic staged trees coupled with the simplification process. Lastly, it can be noticed that the greedy algorithm without a fixed order is fast and comparable to the marginal and total algorithms, despite having to consider multiple variable orderings.

\section{Applications}
\label{sec:app}
\subsection{The Bank Marketing Dataset}
Our first application studies a subset of the \texttt{bank} datset from the UCI machine learning repository reporting information on direct marketing campaigns of a Portuguese banking institution. The marketing campaigns were based on phone calls with the purpose of making customers subscribe to a bank term deposit.

For illustrative purposes we consider only four binary variables: marital status (M = single/married); education (E = uni/no\_uni); contact (C = mobile/landline); subscription (S = yes/no). For each customer, we consider only the first time they were contacted by the bank.

\begin{figure}
\centering
\includegraphics[scale=0.8]{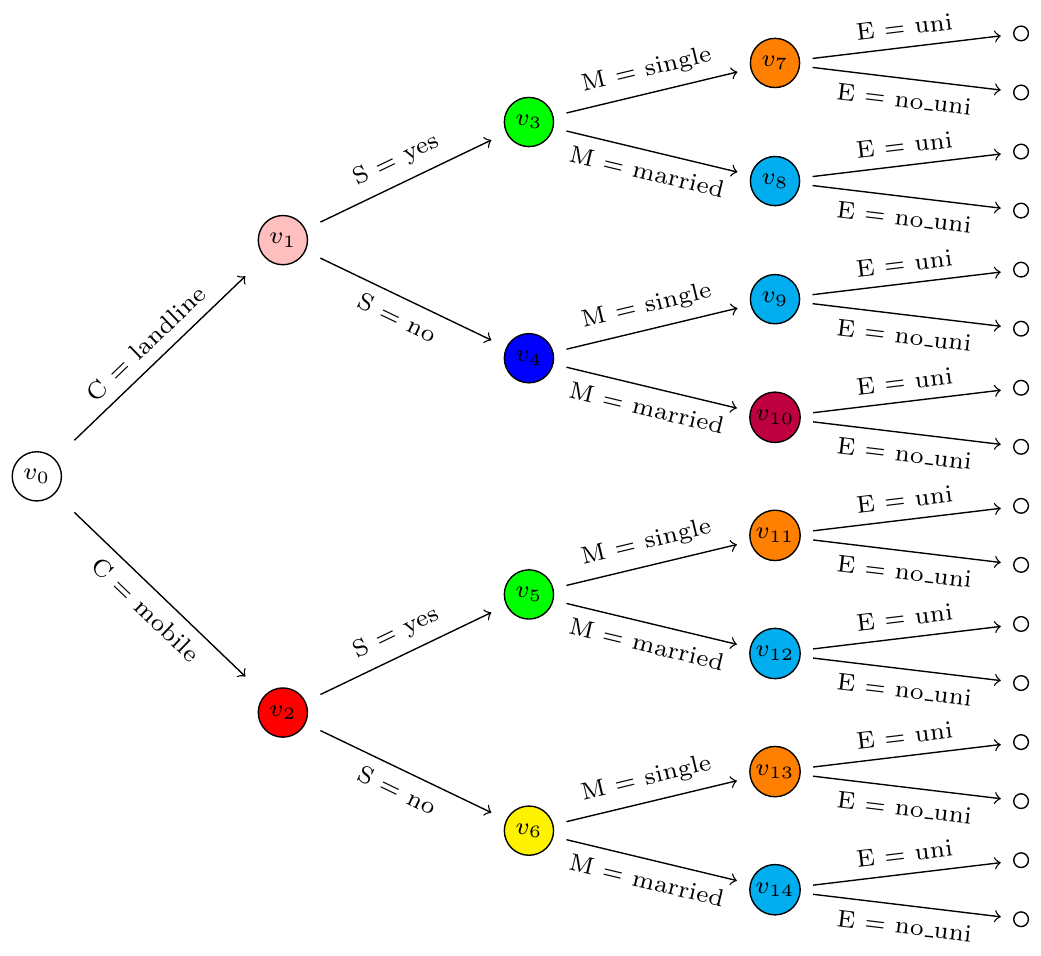}
\caption{Example of an $\bf X$-compatible  staged tree over four binary random variables. \label{fig:stagedbank}}
\end{figure}

The best scoring BN embeds one conditional independence: $\textnormal{S} \independent \textnormal{M} \vert \textnormal{E}, \textnormal{C}$. Whether a customer subscribes to the bank product is independent of her marital status given her educational level and the type of contact. Simple staged trees are learned with all algorithms considered so far and it can be shown that they all outperform the BN in terms of model fit. Here, we consider the simple staged tree learned considering all possible orderings and the marginal algorithm. This is reported in Figure \ref{fig:stagedbank} and it has variable ordering $(\textnormal{C},\textnormal{S},\textnormal{M},\textnormal{E})$. From the staging $\{v_3,v_5\}$ it follows that $\textnormal{M}\independent \textnormal{C}\vert \textnormal{S = yes}$. The staging over the vertices $v_{11},\dots,v_{14}$ implies that $\textnormal{E} \independent \textnormal{S} \vert \textnormal{M}, \textnormal{C = mobile}$. The staging over the vertices $v_{7},\dots,v_{9}$ implies additional highly non-symmetric independences which in general cannot be depicted by BNs.

\begin{figure}
    \centering
    \includegraphics[]{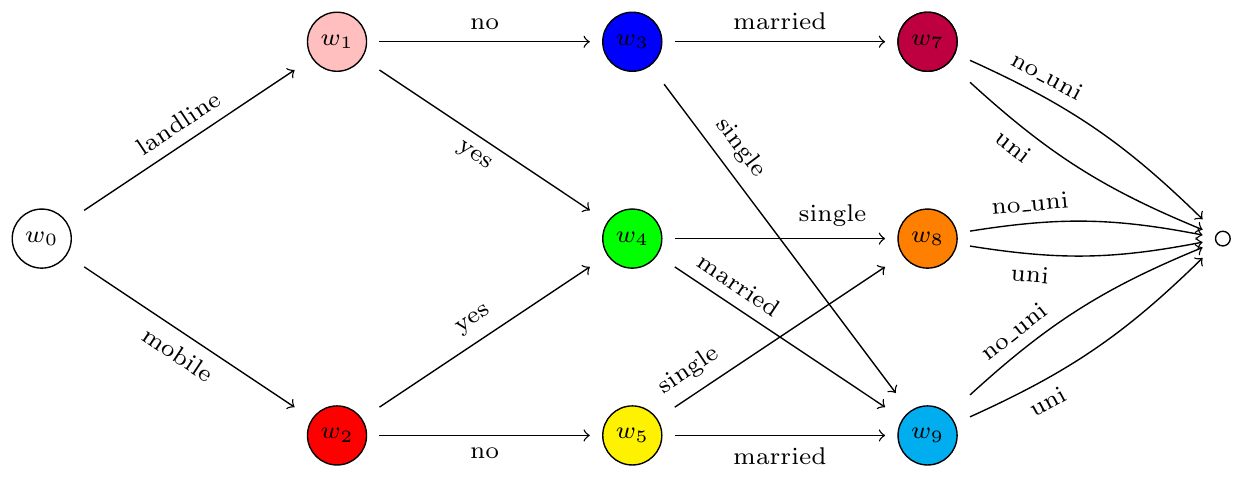}
    \caption{The CEG equivalent to the staged tree in Figure \ref{fig:stagedbank} for the bank marketing dataset.}
    \label{fig:cegbank}
\end{figure}

The equivalent CEG is reported in Figure \ref{fig:cegbank}. Since the staged tree is simple each stage exactly corresponds to a vertex in the CEG, which are colored as in the tree in Figure \ref{fig:stagedbank} for ease of understanding. There are two paths that lead to the position $w_4$ coinciding with a subscriber to the bank product for both landline and mobile contact. This represents the already noticed context-specific independence $\textnormal{M}\independent \textnormal{C}\vert \textnormal{S = yes}$. Furthermore, it also implies that the conditional distribution of subsequent random variables, in this case E, is the same for $(\textnormal{C = mobile}, \textnormal{S = yes})$ and $(\textnormal{C = landline}, \textnormal{S = yes})$. Let's consider position $w_9$: there are four paths leading to this vertex which represent contexts for the variables C, S and M - namely (landline, no, single), (landline, yes, married), (mobile, yes, married) and (mobile, no, married). The conditional distribution of education is the same for these four contexts.

\begin{figure}
\centering
\includegraphics[]{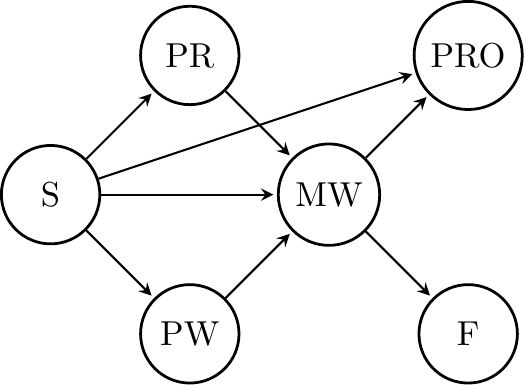}
 \caption{BN learned with hill-climbing for the coronary dataset.\label{fig:bncor}}
 \end{figure}

Although for this application the sample space is small enough that we can read conditional independence directly from the staged tree, it is apparent that the CEG already provides a more compact representation of the model. The compactness of the learned CEG is maximized since the equivalent staged tree has been learned from data to be simple.

\subsection{The Coronary Dataset}

\begin{figure}
    \centering
    \includegraphics[scale=0.9]{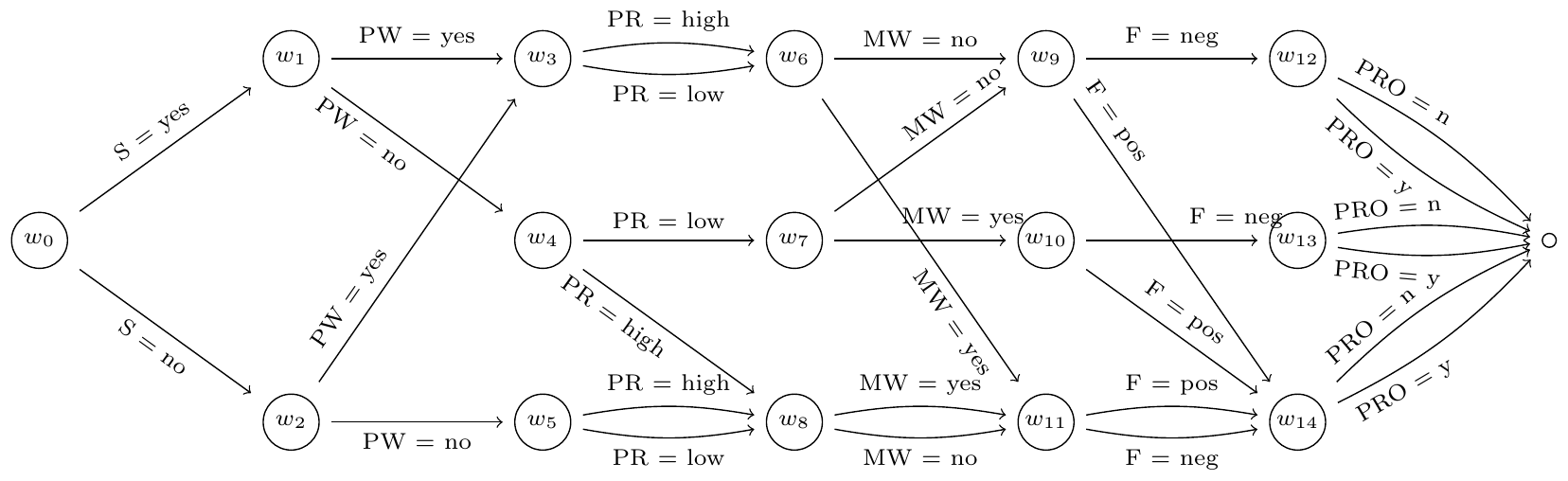}
    \caption{The CEG equivalent to the simple staged tree learned over the coronary dataset with the total algorithm.}\label{fig:cegcor}
\end{figure}

Next we consider the coronary dataset included in \texttt{bnlearn} reporting risk factors of coronary thrombosis of 1841 men. The dataset includes six binary variables: smoking (S=yes/no); strenuous mental work (MW=yes/no); strenuous physical work (PW=yes/no); systolic blood pressure (PR=high/low); family anamnesis of coronary heart disease (F=pos/neg); ratio of beta and alpha lipoproteins (PRO=y/n). The BN model learned with hill-climbing is reported in Figure \ref{fig:bncor} and embeds three conditional independence statements: $\text{PR}\independent \text{PW}\vert\text{S}$, $\text{F}\independent \text{S}, \text{PR}, \text{PW}\vert\text{MW}$ and $\text{PRO}\independent \text{F},\text{PR},\text{PW}\vert\text{MW},\text{S}$.

The simple staged tree learned with the total algorithm provides a better fit to the data (see Table \ref{table:bic}). However, it becomes challenging to visualize it since it has 63 internal nodes and 64 leaves. Conversely, the equivalent CEG has 15 internal nodes and is reported in Figure \ref{fig:cegcor}. Since the equivalent staged tree is simple, each stage of the tree coincides with a vertex of the CEG. Notice that the CEG equivalent to the staged tree learned with backward hill-climbing would have more than 30 vertices (see Table \ref{table:pos}) and would therefore be very challenging to interpret.

The CEG in Figure \ref{fig:cegcor} embeds a wide array of non-symmetric independences that BNs cannot graphically report. For instance the edges $(w_1,w_3)$ and $(w_2,w_3)$ are associated to the context-specific independence $\text{PR}\independent \text{S}\vert\text{PW}=\text{yes}$. The edges between $w_3$ and $w_6$ again represent the context-specific independence $\text{MW}\independent \text{S},\text{PR}\vert\text{PW}=\text{yes}$ and the edges between $w_5$ and $w_8$ are associated to $\text{MW}\independent \text{PR}\vert\text{S}=\text{no},\text{PW}=\text{no}$. The edge $(w_4,w_8)$ can be described as what is usually termed a local independence: the conditional distribution of MW given S = no and PW = no is equal to the conditional distribution of MW given S = yes, PW = no and PR = high. Lastly, we can notice that many edges lead to the position $w_{14}$. This tells us that the conditional distribution of PRO is equal given any conditioning of the preceding variables which can be read from the labels of edges in paths that end in $w_{14}$. This further implies that given F = pos, PRO is independent of all other variables.

\section{Conclusions}

Staged trees and CEGs provide a flexible framework to graphically represent any non-symmetric independence existing in a random vector. However, for generic data-learned staged trees the conversion of the staged tree into the equivalent CEG representation does not provide a significant simplification of the underlying graph. In this paper, we introduced novel structural learning algorithms for the class of simple staged trees for which the conversion into a CEG drastically reduces the number of vertices. Although the requirement of learning a simple staged tree greatly reduces the size of the model search, our experiment demonstrated that simple staged trees often outperform BNs in model fit. The insights that the CEG provides into the dependence structure have been highlighted using two datasets.

We thus claim that if the objective is to learn CEG, the appropriate search space to explore is the one of simple staged tree, otherwise CEG graphical representation can, in general be avoided, since it will not represent a more compact graph with respect to the staged tree.

The algorithms proposed here only search the space of simple staged trees. One possible extension is to consider searching algorithms over the whole model space of staged tree which strongly penalize staging structures leading to CEGs with many vertices. The learned staged tree would not be necessarily simple but still such that the equivalent CEG provides a significant compact representation of the model.

\section*{Acknowledgments}
Gherardo Varando's work was funded by the European Research 
Council (ERC) Synergy Grant “Understanding and Modelling the Earth 
System with Machine Learning (USMILE)” under Grant Agreement No 855187. 

\bibliographystyle{chicago}

\bibliography{refs}

\end{document}